\documentclass[journal=jacsat,manuscript=article]{achemso}
\usepackage[version=3]{mhchem} 
\usepackage{siunitx}
\usepackage[dvipsnames]{xcolor}
\usepackage{amssymb}
\usepackage{pifont}
\usepackage{wrapfig}
\usepackage{hyperref}
\hypersetup{hidelinks}

\usepackage{amsmath} 

\usepackage{tabularx,colortbl}
\usepackage{capt-of}
\usepackage{caption}
\usepackage{subcaption}
\usepackage[font=normalsize,labelfont={bf},font={bf}]{caption}
\author{Aayush Shah}
\email{aayushsh@andrew.cmu.edu}
\affiliation[meche]
{Department of Mechanical Engineering, Carnegie Mellon University, 15213, USA}

\author{Chakradhar Guntuboina}
\affiliation[ece]
{Department of Electrical and Computer Engineering, Carnegie Mellon University, 15213, USA}

\author{Amir Barati Farimani}
\email{barati@cmu.edu}
\affiliation[meche]
{Department of Mechanical Engineering, Carnegie Mellon University, 15213, USA}
\alsoaffiliation[biomed]
{Department of Biomedical Engineering, Carnegie Mellon University, 15213, USA}
\alsoaffiliation[mld]
{Machine Learning Department, Carnegie Mellon University, 15213, USA}

\title[An \textsf{achemso} demo]
{Peptide-GPT: Generative Design of Peptides using Generative Pre-trained Transformers and Bio-informatic Supervision}

\begin{document}

\begin{abstract}


In recent years, natural language processing (NLP) models have demonstrated remarkable capabilities in various domains beyond traditional text generation. In this work, we introduce PeptideGPT, a protein language model tailored to generate protein sequences with distinct properties: hemolytic activity, solubility, and non-fouling characteristics. 
To facilitate a rigorous evaluation of these generated sequences, we established a comprehensive evaluation pipeline consisting of ideas from bioinformatics to retain valid proteins with ordered structures. First, we rank the generated sequences based on their perplexity scores, then we filter out those lying outside the permissible convex hull of proteins. Finally, we predict the structure using ESMFold and select the proteins with pLDDT values greater than 70 to ensure ordered structure. 


The properties of generated sequences are evaluated using task-specific classifiers - PeptideBERT and HAPPENN.
We achieved an accuracy of 76.26\% in hemolytic, 72.46\% in non-hemolytic, 78.84\% in non-fouling, and 68.06\% in solubility protein generation. 
Our experimental results demonstrate the effectiveness of PeptideGPT in de novo protein design and underscore the potential of leveraging NLP-based approaches for paving the way for future innovations and breakthroughs in synthetic biology and bioinformatics.

Codes, models, and data used in this study are freely available at: 

\url{https://github.com/aayush-shah14/PeptideGPT}

\end{abstract}

\section{Introduction}

Advances in deep learning, particularly natural language processing, have given rise to powerful large language models that are capable of generating human-like text by understanding the underlying sentiments and contexts.\cite{NIPS2017_3f5ee243, Radford2018ImprovingLU, brown2020language, zellers2019defending}. Just as the arrangement of letters in words and sentences conveys meaning in human languages, proteins can be described as a sequence of amino acids, with each amino acid having a particular letter assigned to it. The sequence of amino acids in a protein determines its structure, function, and properties\cite{stryer2002biochemistry, anfinsen1973principles}. This analogy between protein sequences and human language allows for natural language processing tools to be useful in protein biology \cite{Qiu2024.04.17.589642, Johnson2024, Shuai2021.12.13.472419, Kim2024}. Protein design refers to the process of creating new proteins with desired structures, functions, or properties using computational, experimental, or a combination of both approaches. This can have numerous practical applications in the fields of drug discovery, targeted therapeutics, and environmental sustainability \cite{ELGAMACY202285, kuan2024abgptnovoantibodydesign, Cao2024}. Recent works have explored the usage of generative large language models for enabling protein design. ProGEN utilizes a 1.2B language model to enable controllable generation of proteins given a molecular function or cellular component \cite{Madani2023}. ProtGPT2 is a language model based on GPT 2 \cite{Radford2019LanguageMA} which has been trained on the protein space to generative novel proteins which follow the principle of natural ones \cite{ferruz2022proteingpt2}. ProteinMPNN aims to find an amino acid sequence that will fold to a given a protein backbone structure of interest \cite{Dauparas2022.06.03.494563}. RFDiffusion uses denoising diffusion probabilistic models (DDPM) to achieve unconditional and topology-constrained protein monomer design, protein binder design, symmetric oligomer design, enzyme active site scaffolding, and symmetric motif scaffolding for therapeutic and metal-binding protein design \cite{Watson2023}.

In this work, we focus on producing proteins having a specific essential peptide property. We have created a set of models, collectively termed as PeptideGPT, each of which is capable of producing proteins having a certain property, namely, hemolysis, non-hemolysis, non-fouling, and solubility. These properties are defined by the sequence of amino acids \cite{Dunn2015, Varanko2020, Schueler-Furman2017}. 

Hemolytic proteins are substances that can cause the lysis (or rupture) of red blood cells, leading to the release of hemoglobin. Such proteins have applications in medicine, particularly in the treatment of certain conditions where targeted cell lysis is desired. Generation of non-hemolytic proteins is also desired as they can be used as therapeutic agents in medicine \cite{plisson2020machine}. Non-fouling proteins are proteins that resist the nonspecific adsorption or binding of other substances onto their surfaces. This is important in medical implants and drug delivery systems as they can help prevent the attachment of proteins and cells that could lead to biofilm formation, thrombosis, inflammation, and infection \cite{harding2014combating, yu2011anti}. Soluble proteins are essential for structural biology and biochemical assays since insoluble proteins can aggregate and may lose their native conformation, activity, and functionality, making them unsuitable for reliable experimentation. Thus, the creation of soluble proteins is desired for drug development and bio-molecular engineering \cite{sarma2018peptide}. 

PeptideGPT has been developed by fine-tuning ProtGPT2 on datasets consisting solely of hemolytic, non-hemolytic, non-fouling, and soluble proteins. The sequences generated by PeptideGPT undergo a series of filtering steps rooted in bioinformatics principles, comprising of a protein validity check and structure evaluation. 


We then predict the property of the sequences after the final stage of the evaluation using a classifier dedicated for that task. The accuracy given by this prediction stage be inferred as the percentage of valid proteins with an ordered good structure having the desired property. PeptideGPT scores an accuracy of 76.26\% on hemolytic, 72.46\% on non-hemolytic, 78.84\% on non-fouling and 68.06\% on soluble protein generation tasks. This shows that PeptideGPT is able to capture the relationship between protein sequence and its property to generate sequences accordingly. We now lay out our methodology for obtaining the datasets followed by an explanation of each stage in our inference pipeline.
\vspace{2mm}

\begin{figure}
    \centering
    \includegraphics[width=\textwidth]{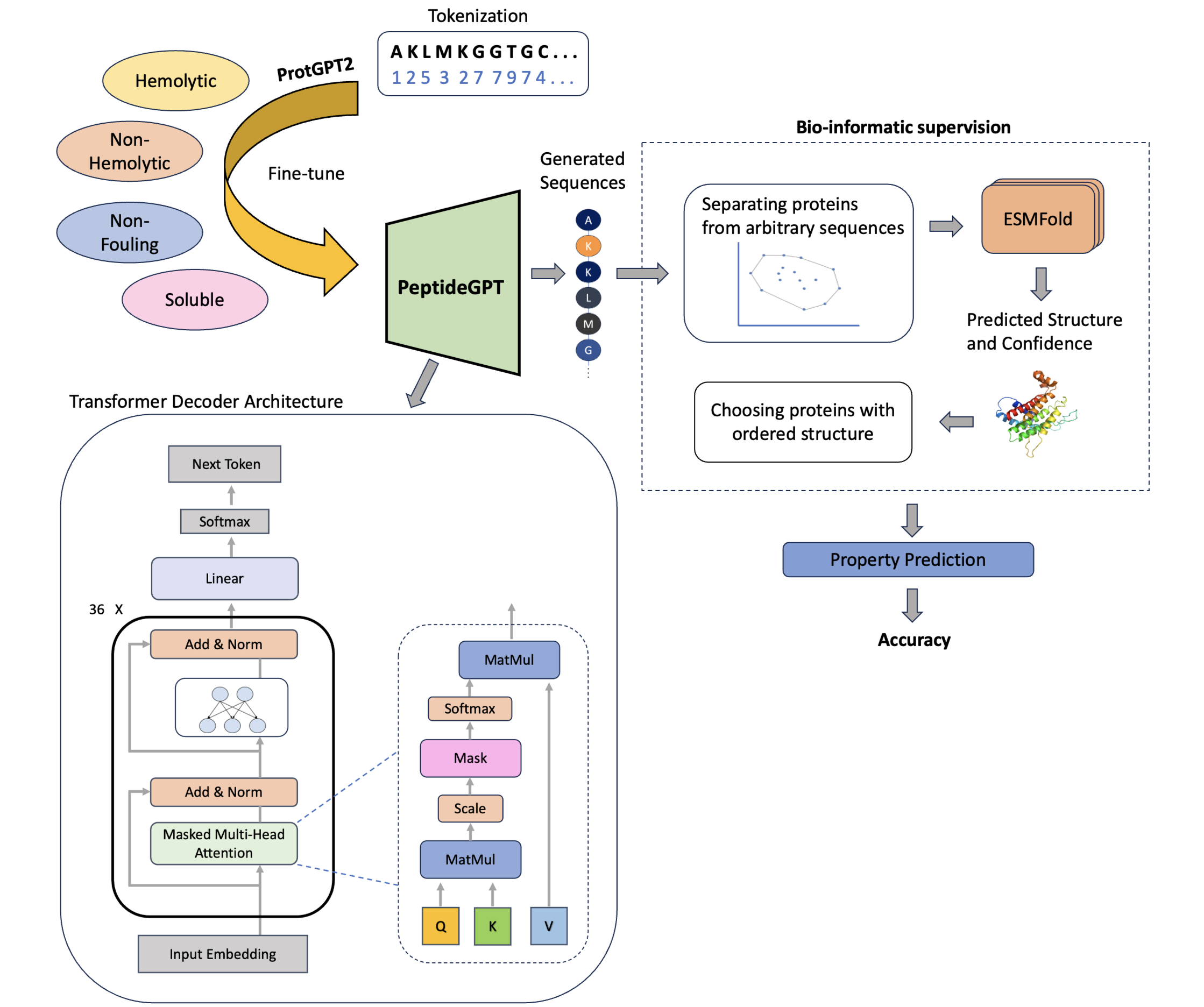}
    \vspace{2mm}
    \caption{The pipeline for PeptideGPT. The peptide sequences are tokenized and passed through ProtGPT2 for fine-tuning on the four tasks. The generated peptides undergo bio-informatic supervision to ensure that only valid proteins with an ordered structure are chosen as the final output. The property is then evaluated through a suitable classifier.}
    \label{fig:pipeline}
\end{figure}

\section{Methodology}

\subsection{Dataset Preparation}

For each task, we curated a dataset consisting solely of the proteins possessing the property we aim to confer onto the generated sequences. Experimentally verified hemolytic protein sequences were taken from the Database of Antimicrobial Activity and Structure of Peptides (DBAASPv3) and Hemolytik dataset \cite{gautam2014hemolytik} and the non-hemolytic protein sequences were procured from DBAASPv3 as well. The dataset for non-fouling proteins was generated following Barrett et al \cite{https://doi.org/10.1002/pep2.24079}. For creating the dataset containing soluble proteins, we used information sourced from PROSO II \cite{Smialowski2012}, where the solubility of the sequences was determined through a retrospective evaluation of electronic laboratory notebooks. In each of these datasets, duplicate entries were removed and sequences were modified according to the input format for ProtGPT2 by adding a $<|endoftext|>$ token at the start and end of each sequence and introducing a newline character after every 60 amino acids. The details of the datasets are presented in Table \ref{data}. 

\begin{table}[H]
\centering
\begin{tabular}{lcccc}
\hline
\textbf{} & \textbf{Hemolytic} & \textbf{Non-Hemolytic} & \textbf{Non-fouling} & \textbf{Soluble} \\ \hline
\textbf{No. Train data} & 1487 & 6741 & 2880 & 7906 \\
\textbf{No. Validation data} & 165 & 749 & 720 & 879 \\
\textbf{Avg Sequence Length} & 19 & 19 & 6 & 136 \\ 
\hline
\end{tabular}
\caption{Summary of datasets}
\label{data}
\end{table}

\begin{figure}[h!]
  \centering
  \begin{subfigure}[b]{0.45\linewidth}
    \includegraphics[width=\linewidth]{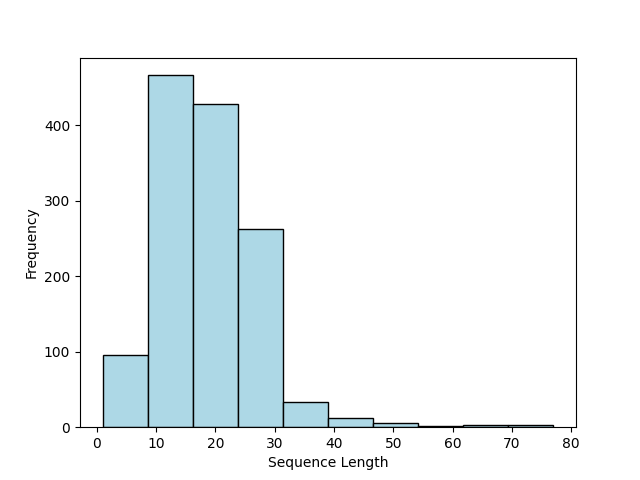}
    \caption{}
  \end{subfigure}
  \hfill
  \begin{subfigure}[b]{0.45\linewidth}
    \includegraphics[width=\linewidth]{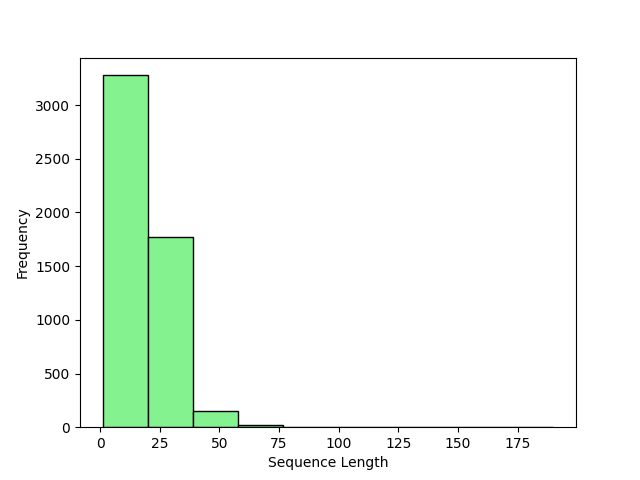}
    \caption{}
  \end{subfigure}
  \vskip\baselineskip
  \begin{subfigure}[b]{0.45\linewidth}
    \includegraphics[width=\linewidth]{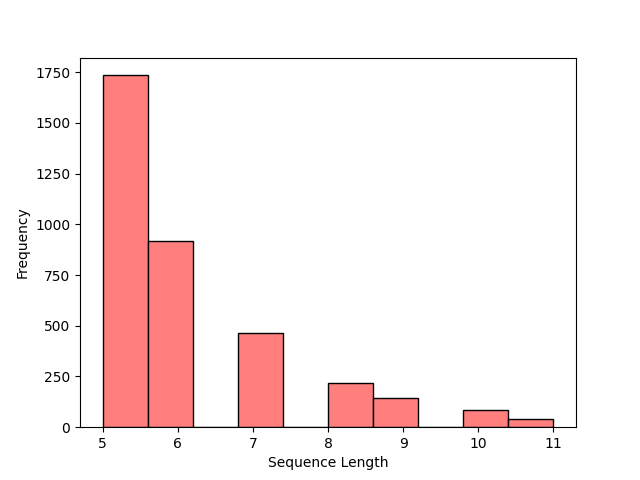}
    \caption{}
  \end{subfigure}
  \hfill
  \begin{subfigure}[b]{0.45\linewidth}
    \includegraphics[width=\linewidth]{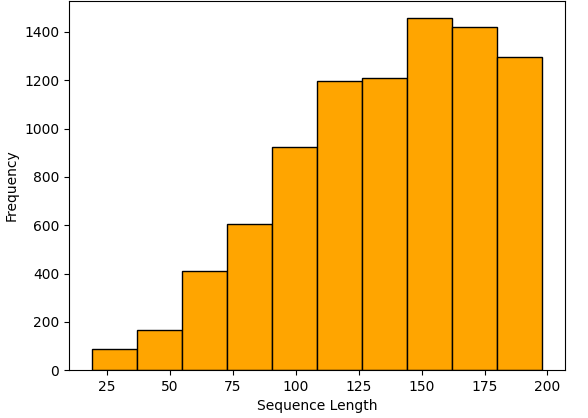}
    \caption{}
  \end{subfigure}
  
  \caption{Sequence length distribution of each Peptide property dataset (a) Hemolytic, (b) Non-Hemolytic, (c) Non-Fouling, and (d) Solubility.}
  \label{fig:seqlens}
\end{figure}

The datasets employed for each specific task and their corresponding sequence length distributions are visually depicted in Figure \ref{fig:seqlens}. Each dataset consists only of those proteins which have the property under consideration. The dataset focusing on non-fouling properties comprises sequences ranging from 2 to 20 residues, emphasizing shorter sequences to capture specific characteristics pertinent to non-fouling. In contrast, the dataset for solubility includes sequences with lengths spanning from 18 to 198 residues, indicating a diverse range accommodating various structural and functional attributes. This varied sequence length distribution requires the tailoring of generation parameters according to the task.

\subsection{Model Architecture}

The ProtGPT2 architecture was employed as the foundational model for our approach which was downloaded from HuggingFace. The architecture follows GPT2-large Transformer, featuring 36 layers and a model dimensionality of 1280. A multi-headed attention scheme is implemented with 20 attention heads, resulting in a head dimensionality of 64. The model has a parameter count of 738 million which were initialized from ProtGPT2's pre-trained weights. For tokenizing the inputs, a Byte-Pair Encoding (BPE) tokenizer was utilized, trained on over 0.5 million sequences sourced from Swiss-Prot. The vocabulary encompasses 50,256 tokens, with an average token size of four amino acids. The maximum input size for the model is 1024 tokens.

\subsection{Training Procedure}

A separate model was fine-tuned on the corresponding dataset for each task following a causal language modeling strategy, where the model predicts the next token in a sequence based on the preceding tokens. In other words, the model is conditioned on the context provided by the previous tokens to generate the subsequent token. This approach is autoregressive, where the loss function employed is cross-entropy between the output of the model and the ground truth, which is just the inputs shifted by one token to the left.

\begin{equation}
    \mathcal{L}_{\text{CLM}} = - \sum_{k=1}^{D} \log p_{\theta} \left( w_{i}^{k} \,|\, w_{<i}^{k} \right),
    \label{eq:clm_loss}
\end{equation}

In Equation \ref{eq:clm_loss}, $D$ is the size of the dataset, $w_{i}^{k}$ represents the $i$-th token in the $k$-th sequence, and $w_{<i}^{k}$ denotes the preceding context.

 The training process was stopped once the loss curves converged. The model was trained on a single NVIDIA RTX A6000 GPU with a memory of 48 GB. 
 
\subsection{Sampling Procedure}
The generation process from the trained model is influenced by certain key parameters, namely, repetition penalty, $top_k$, maximum sequence length, and number of generated sequences. The repetition penalty is a parameter which discourages the repetition of tokens within generated sequences. The $top_k$ parameter specifies the sampling process for the next token, where k top-most probable tokens are considered at each step of text generation. We have kept the repetition penalty as 1.2 and a $top_k$ value of 950. Both of these values were found to be the best-performing parameters by Ferruz et al\cite{ferruz2022proteingpt2}. The maximum sequence length specifies the maximum number of tokens in the generated outputs. Since the average number of amino acids per token was four, we kept this parameter according to the maximum length observed in Figure \ref{fig:seqlens}. After generation, we ranked the outputs according to their perplexity. Perplexity is the exponentiation of the entropy of the probability distribution over the predicted tokens and is a measure of the model's confidence in its generation. We selected the top one-third of these sequences for further processing to ensure that the best outputs are chosen. 

\subsection{Protein Verification}
\begin{figure}
    \centering
    \begin{subfigure}[b]{0.4\textwidth}
        \centering
        \includegraphics[width=1.3\textwidth]{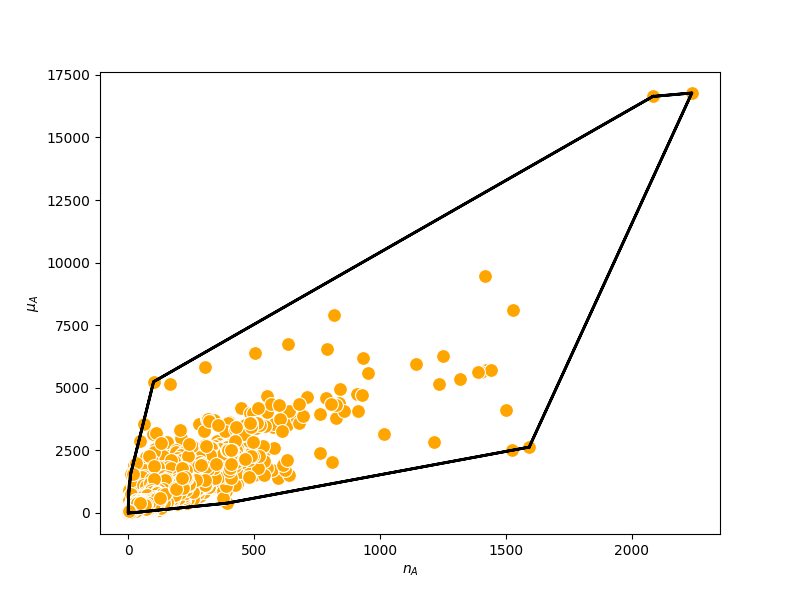}
        \caption{$n_A$ vs $\mu_A$}
        \label{fig:sub1}
    \end{subfigure}
    \hfill
    \begin{subfigure}[b]{0.4\textwidth}
        \centering
        \includegraphics[width=1.3\textwidth]{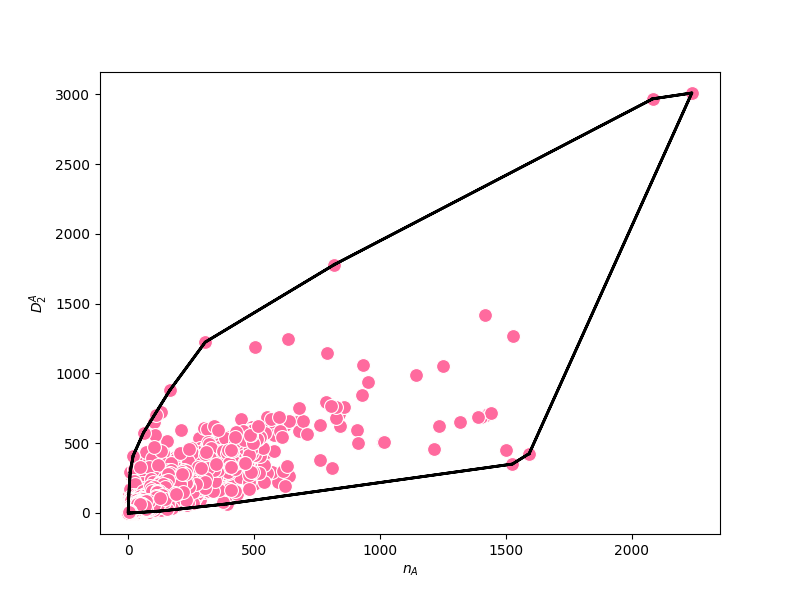}
        \caption{$n_A$ vs $D_2^A$}
        \label{fig:sub2}
    \end{subfigure}
    \caption{Convex hulls for Alanine. Points inside these hulls are considered to be valid protein sequences.}
    \label{fig:both}
\end{figure}

Although our model is intended to generate proteins, the generated sequence might not always conform to the rules pertaining to an amino acid sequence being a protein. This is due to the inherently generative nature of Transformers and the complexities of protein sequence prediction. This can have unwanted implications in certain downstream tasks which might require the synthesis of generated sequences. Hence, we employed a bioinformatics technique which distinguishes proteins from arbitrary amino acid sequences \cite{Yau2015}. Each sequence is represented as a point with three quantities as coordinates: $n_k$ the number of occurrences of the amino acid “k” within the sequence, $\mu_k$ the mean distance of the amino acid “k” from the first position and $D^2_k$ the second normalized central moment of the distribution of amino acid “k”. Yau et al. found that the points corresponding to proteins are concentrated in a convex hull out of the entire amino acid space with a 99.69 \% accuracy. Following their work, we created 20 such convex hulls for each amino acid from 571,282 proteins belonging to the reviewed (Swiss-Prot) part of the 2024\_02 release of UniProtKB. An example of the convex hulls for the amino acid Alanine is shown in Figures \ref{fig:sub1} and \ref{fig:sub2}. From the sequences generated by our model, we only selected those which were present inside all of the 20 convex hulls. Around 6.5\% of the generated peptides are eliminated in this manner, ensuring only valid proteins are passed for further evaluation. 


\subsection{Structure Prediction}

An important feature to consider when designing de novo protein sequences is their ability to fold into stable ordered structures. We employed ESMFold, an AI model that predicts protein structure from a sequence of genes \cite{Lin2022.07.20.500902}. ESMFold is built on a 15B parameter Transformer model and achieves accuracy comparable to other state-of-the-art models with an order-of-magnitude inference time speedup. ESMFold evaluates the three-dimensional structures predicted from the generated sequences and produces a pLDDT metric for the sequence. pLDDT is a proxy for structural order and stands for Predicted Local Distance Difference Test. It corresponds to the model’s predicted score on the lDDT-C$\alpha$ metric. Regions with pLDDT between 70 and 90 are expected to be modeled well and have an ordered structure \cite{jumper2021highly, varadi2021alphafold} whereas low scores (pLDDT less than 50) tend to appear in disordered regions \cite{Tunyasuvunakool2021}. 
ESMFold has a faster inference time over other structure prediction models like AlphaFold \cite{jumper2021highly} and ColabFold \cite{Mirdita2022}, which is important since the main bottleneck in the inference pipeline is the structure prediction stage. 
We picked sequences with pLDDT values above 70 as the final output of our generation process. These checks during inference have been implemented to ensure a higher likelihood of yielding final outputs consisting of valid proteins with ordered structures. 

\subsection{Property Classification}



The protein sequences that pass the previous stages are evaluated to determine whether they indeed possess the desired properties. To accomplish this, we employed specialized prediction models tailored to each specific property of interest.

For predicting the hemolytic property, we utilized the HAPPENN model, a neural network-based approach designed for hemolytic activity prediction\cite{timmons2020happenn}.

For predicting non-fouling and solubility properties, we used PeptideBERT, a state-of-the-art peptide encoder built on the BERT architecture. PeptideBERT incorporates 12 attention heads and 12 hidden layers, enabling it to capture relationships between protein sequences and their property \cite{guntuboina2023peptidebert}. Specifically for the non-fouling task, we observed that PeptideBERT was trained on an unbalanced dataset, resulting in a drop in its accuracy. Hence, we used a version of PeptideBERT after training it on a balanced dataset. 

The performance of our model can be measured through the accuracy obtained after running the property prediction model on a set of generated sequences. This can be inferred as the percentage of valid and structurally sound proteins that exhibit the intended property. 

The entire generation and inference pipeline is shown in Figure \ref{fig:pipeline}.

\section{Results and Discussion}

\begin{figure}
    \centering
    \includegraphics[width=\textwidth]{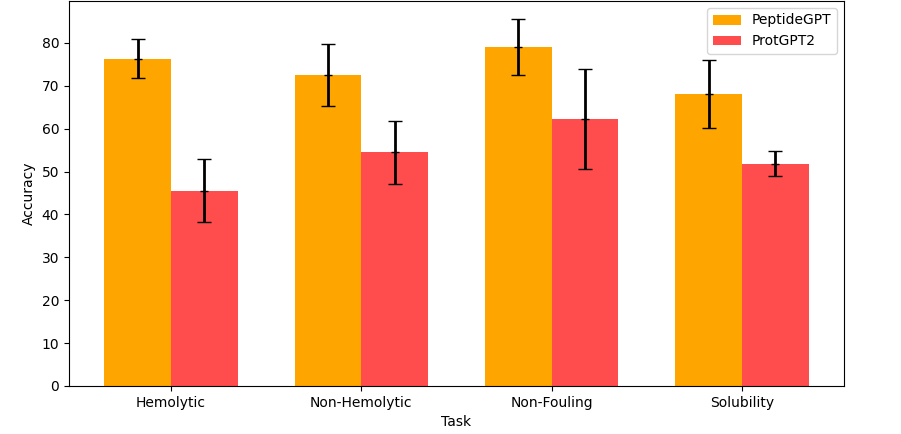}
    \caption{Accuracy of PeptideGPT vs ProtGPT2 generated sequences on each task. We can see an increase in the accuracy due to fine-tuning.}
    \label{fig:accs}
\end{figure}

To demonstrate the effectiveness of PeptideGPT in generating the sequences with the intended property, we predicted their property using existing classifier models for that property - PeptideBERT for non-fouling and solubility and HAPPENN for hemolytic activity. To ensure a robust evaluation, we ran the generation process for five different seeds for each task, each time producing 1000 sequences resulting in a total of 5000 sequences over all the tasks. These were passed through the inference pipeline outlined in the previous section for choosing the sequences which are valid proteins and possess an ordered structure, as quantified through the pLDDT metric. The average accuracy along with their standard deviation obtained for each task is given in Figure \ref{fig:accs}. We compared the accuracy of PeptideGPT with that of ProtGPT2 to highlight the improvement due to fine-tuning. The accuracy corresponding to ProtGPT2 can also be viewed as a proxy for the percentage of naturally occurring proteins having the property under consideration. It can be seen that a significant jump in accuracy is achieved for each task following fine-tuning, proving the effectiveness of our method. We should also note that the PeptideGPT's accuracy will be limited by the accuracy of the classifier used, since the former aims to generate all positive classes (and negative class for non-hemolytic generation) whereas the latter is evaluated on actual proteins known to have that property. It is observed that PeptideGPT's accuracy resembles the accuracy of the classifiers as shown in Table \ref{res}. A more comprehensive evaluation would entail synthesizing the generated proteins and conduct laboratory testing for evaluating the properties, however that is beyond the scope of this work. 

\begin{table}[H]

\begin{tabular}{lccc}
\hline
\textbf{Task} & \textbf{\% PeptideGPT Accuracy} & \textbf{Classifier} & \textbf{\% Accuracy of Classifier} \\
\hline
Hemolytic & 76.26 $\pm$ 4.54 & HAPPENN & 82.85 \\
\hline
Non - Hemolytic & 72.46 $\pm$ 7.19 & HAPPENN & 82.85 \\
\hline
Non - Fouling & 78.84 $\pm$ 6.52 & PeptideBERT & 88.36 \\
\hline
Solubility & 68.06 $\pm$ 7.88 & PeptideBERT & 70.02 \\
\hline
\end{tabular}
\caption{Accuracy of PeptideGPT and the property classifier used}
\label{res}
\end{table}

\begin{figure}[h!]
  \centering
  \begin{subfigure}[b]{0.45\linewidth}
    \includegraphics[width=\linewidth]{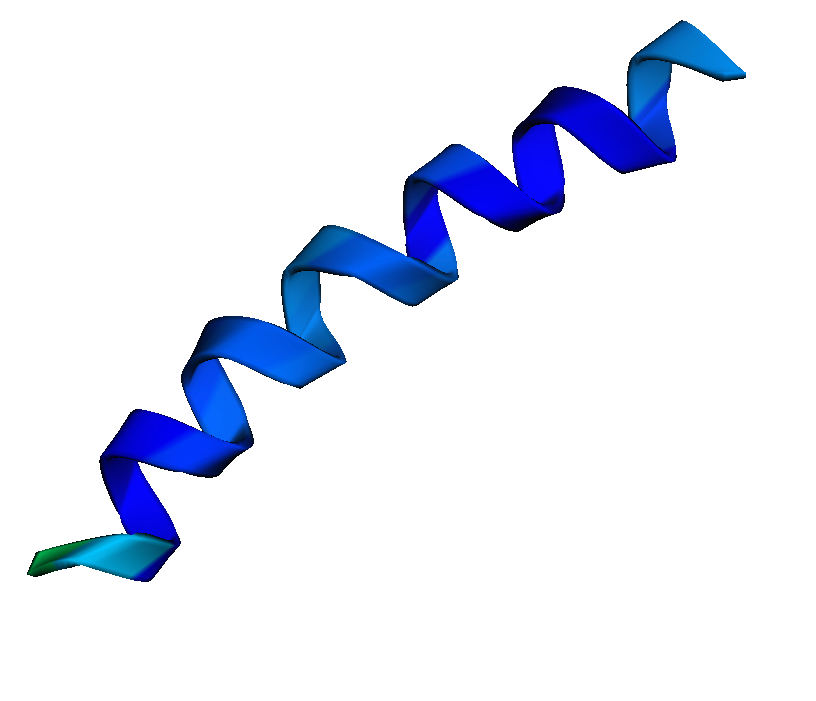}
    \caption{}
  \end{subfigure}
  \hfill
  \begin{subfigure}[b]{0.45\linewidth}
    \includegraphics[width=\linewidth]{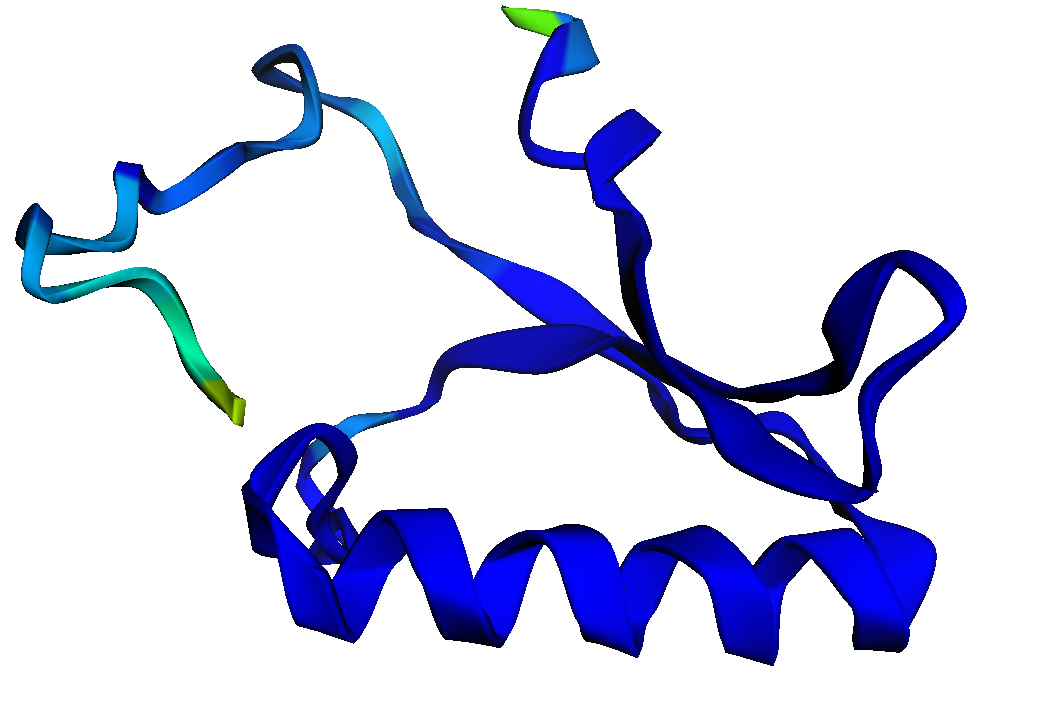}
    \caption{}
  \end{subfigure}
  \vskip\baselineskip
  \begin{subfigure}[b]{0.45\linewidth}
    \includegraphics[width=\linewidth]{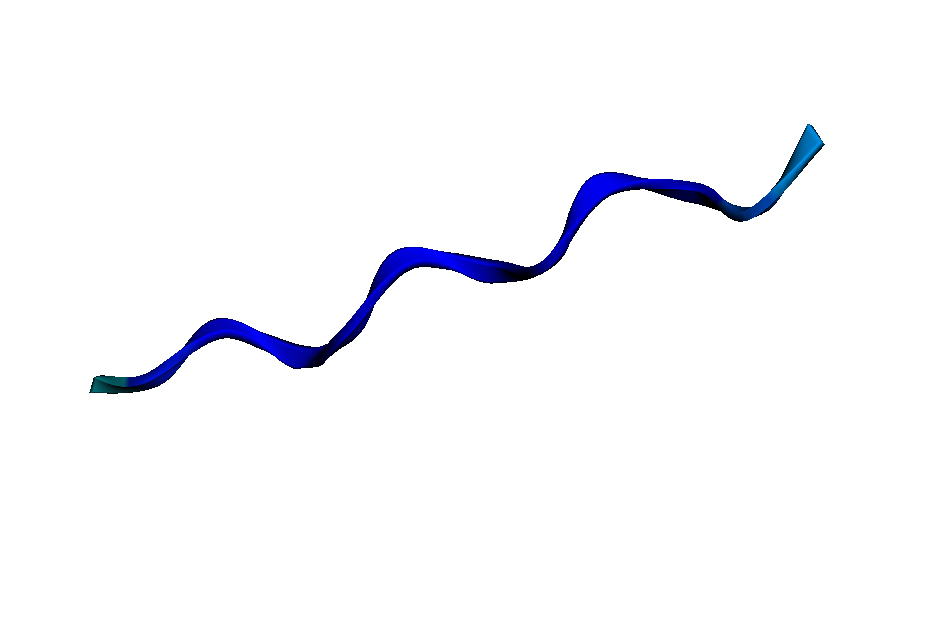}
    \caption{}
  \end{subfigure}
  \hfill
  \begin{subfigure}[b]{0.45\linewidth}
    \includegraphics[width=\linewidth]{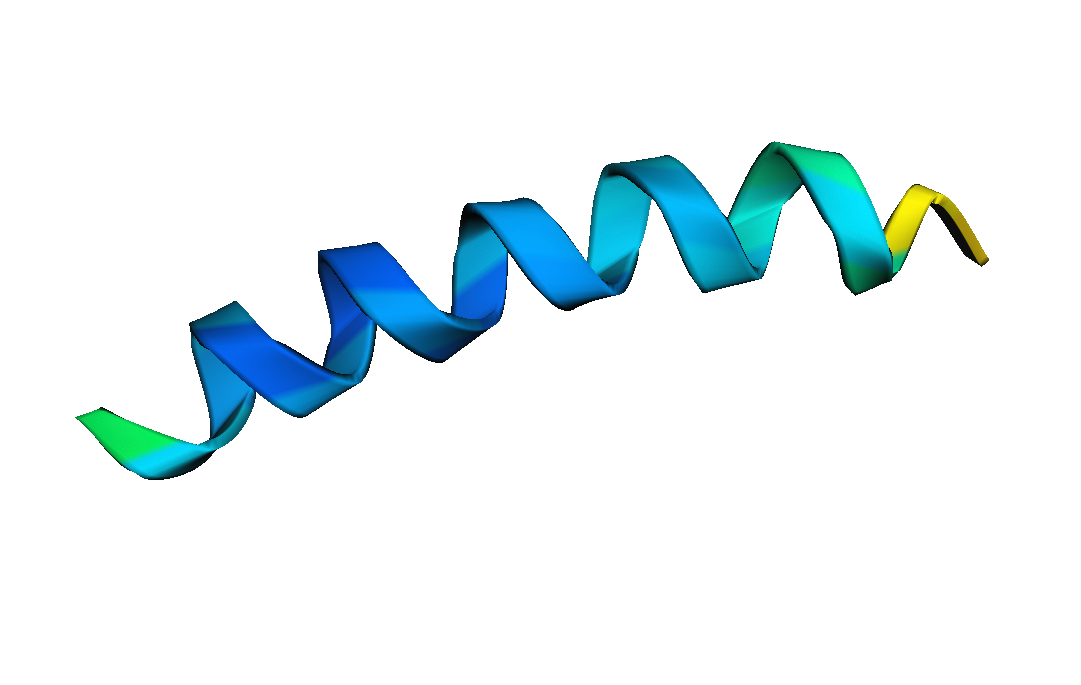}
    \caption{}
  \end{subfigure}
   \vskip\baselineskip
  \begin{subfigure}[b]{\linewidth}
    \centering
    \includegraphics[width=0.8\linewidth]{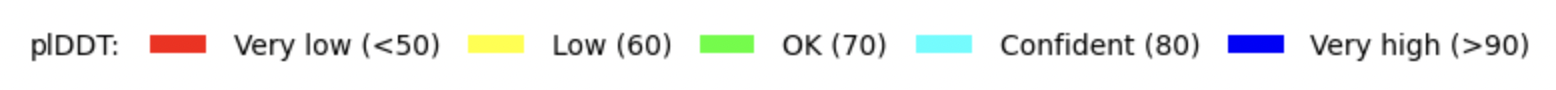}
  \end{subfigure}
  \caption{Example of generated sequences each predicted to have the desired property: a) Hemolytic b) Soluble c) Non-Fouling d) Non-Hemolytic}
  \label{fig:exprots}

\end{figure}

On average, 24\% of total sequences generated by PeptideGPT had a pLDDT value of over 70. Examples of some of the generated sequences for each property are shown in Figure \ref{fig:exprots}. The regions in blue correspond to high pLDDT scores, showing that PeptideGPT indeed generates proteins with ordered and stable structures. 

\vspace{2mm}

Several works have hypothesized that the evolutionary course of nature has not fully explored all possible sets of folded and stable protein sequences\cite{baker2019denovo}. Through the exploration of such uncharted areas, de novo proteins can potentially develop new functions or enhance existing ones. This exploration may lead to proteins with unique or optimized characteristics for specific applications, ranging from industrial enzymes to therapeutic agents. Hence, it is important to see whether sequences generated by PeptideGPT are populating unexplored regions of existing protein spaces. The Transformer's attention mechanism allows each token embedding in the encoder to understand the entire input sequence. However, in practice, the classification token, or the [CLS] token, is typically used as a representation of the entire sequence \cite{schwaller2019molecular, schwaller2020mapping}. PeptideBERT incorporates this [CLS] token at the beginning of each sequence, allowing it to gather information from all other token embeddings into a compact representation. The vector embedding corresponding to the [CLS] token from the final hidden layer from PeptideBERT encoder was then mapped to a two-dimensional space using the t-SNE dimensionality reduction technique.

\begin{figure}[h!]
  \centering
  \begin{subfigure}[b]{0.45\linewidth}
    \includegraphics[width=\linewidth]{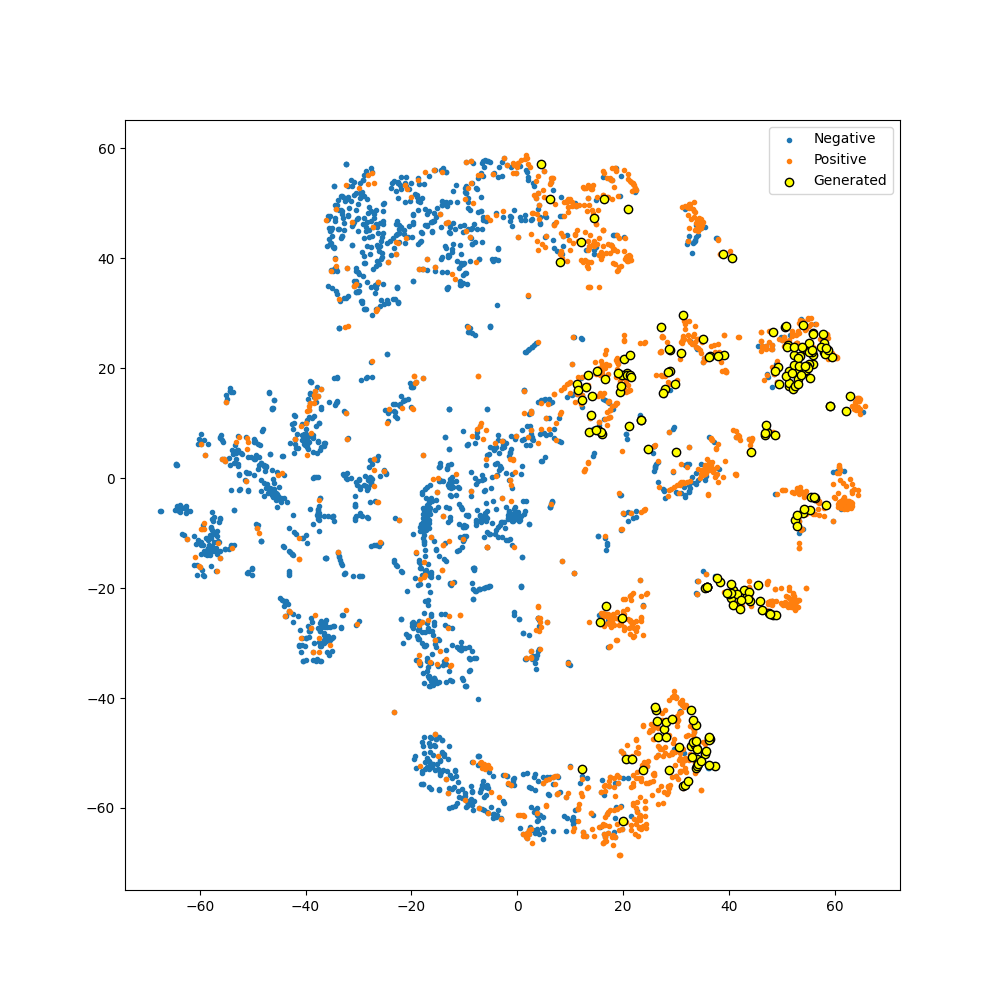}
    \caption{}
  \end{subfigure}
  \hfill
  \begin{subfigure}[b]{0.45\linewidth}
    \includegraphics[width=\linewidth]{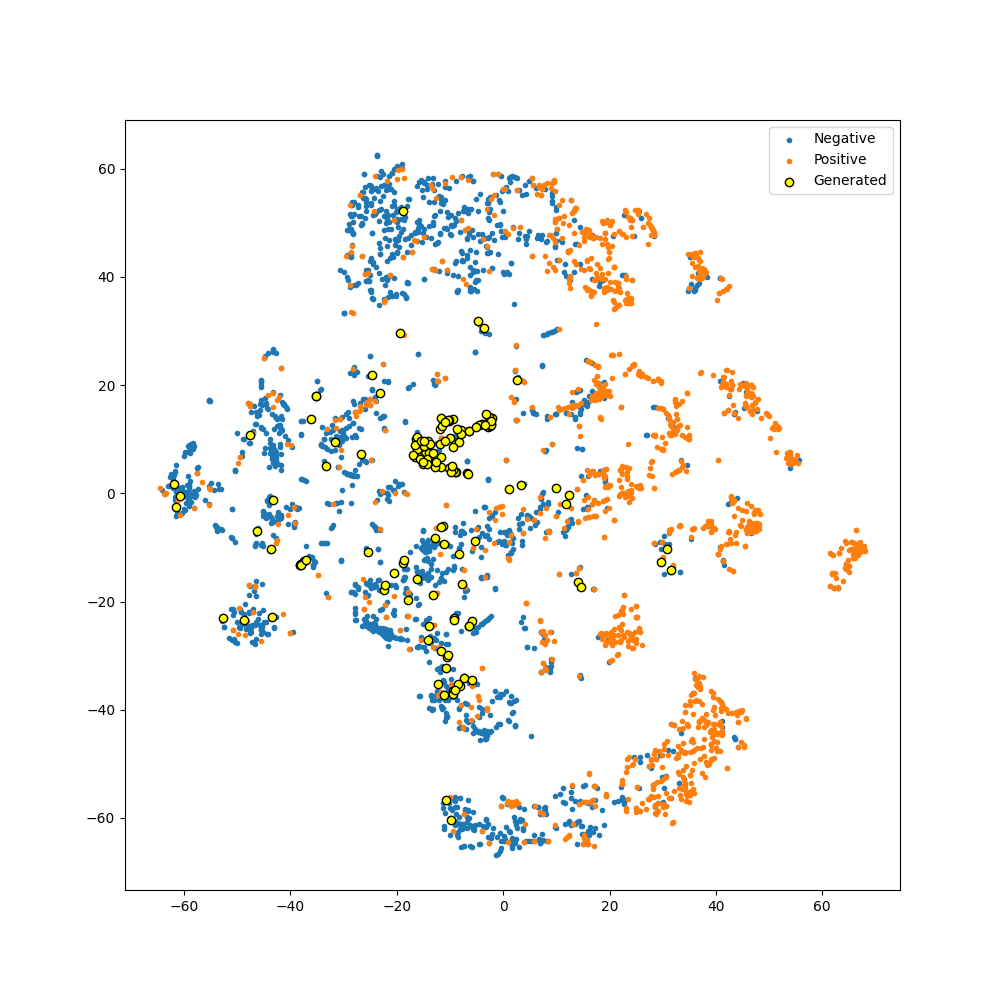}
    \caption{}
  \end{subfigure}
  \vskip\baselineskip
  \begin{subfigure}[b]{0.45\linewidth}
    \includegraphics[width=\linewidth]{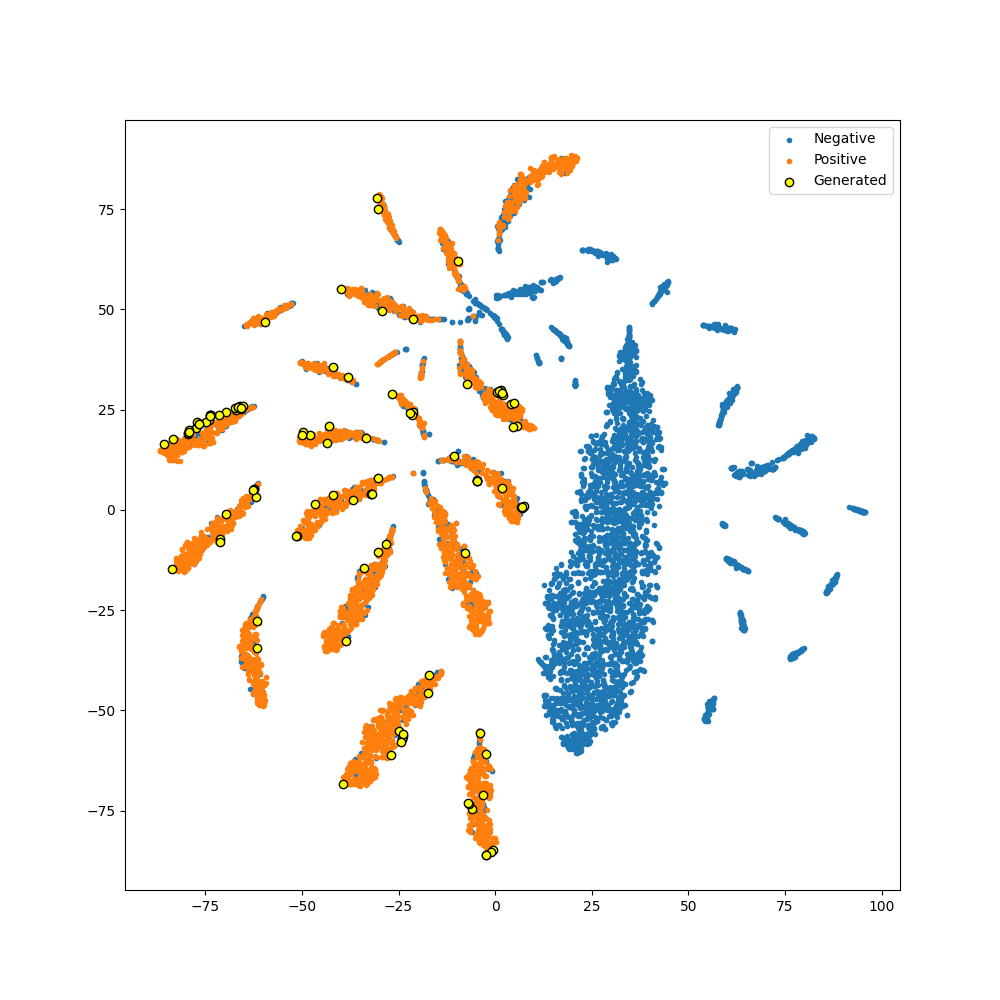}
    \caption{}
  \end{subfigure}
  \hfill
  \begin{subfigure}[b]{0.45\linewidth}
    \includegraphics[width=\linewidth]{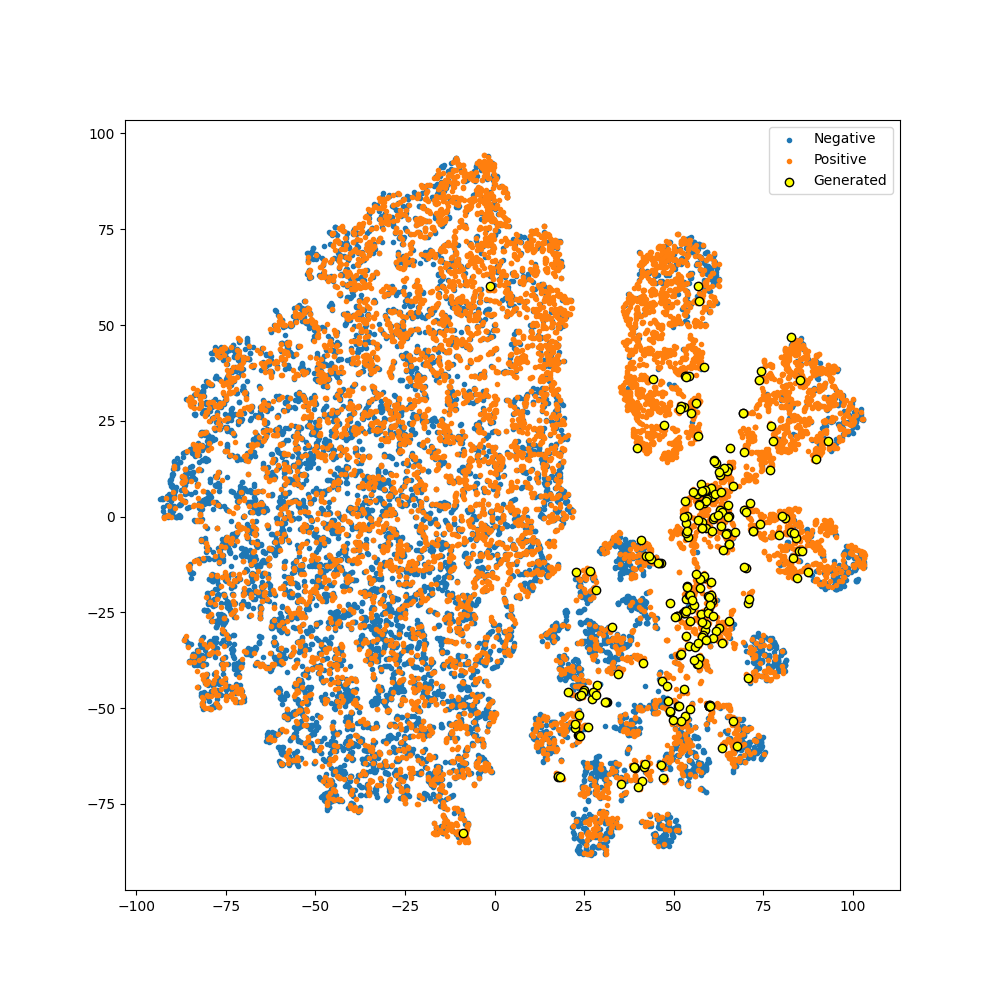}
    \caption{}
  \end{subfigure}
  \caption{t-SNE plots of generated sequences against real sequences for (a) Hemolytic, (b) Non-Hemolytic, (c) Non-Fouling, and (d) Soluble. We can see the generated sequences lie closer to the class whose property they are supposed to inherit.}
  \label{fig:quadrant}
\end{figure}

This was done for three classes of protein sequences - positive (actual proteins which are known to have that property), negative (actual proteins which do not possess the property), and generated (sequences generated by PeptideGPT intended to inherit the given property). The resulting plots are given in Figure \ref{fig:quadrant}. We can observe that the generated sequences tend to align closer to the space occupied by the actual proteins whose properties they are supposed to inherit. Particularly, in the t-SNE plot for the non-fouling task, the positive proteins form island-like regions with the generated sequences situated on their boundary, suggesting that it is on the periphery of known protein space and trying to explore novel regions. This result serves as a promising direction for further in-silico protein design approaches to capture properties and functions not found in naturally occurring proteins.



\section{Conclusion}

In this study, we presented PeptideGPT, an innovative adaptation of generative pre-trained transformers tailored for the generative design of protein sequences with specific functional properties. Our comprehensive assessment, which involved a pipeline combining bioinformatics tools and deep learning models, demonstrated the capability of PeptideGPT to generate protein sequences with high reliability and targeted functionalities such as hemolytic activity, solubility, and non-fouling characteristics.

The results underscore the usefulness of PeptideGPT by achieving accuracies of 76.26\% in hemolytic, 72.46\% in non-hemolytic, 78.84\% in non-fouling, and 68.06\% in solubility tasks, which are significant given the complexity of protein design. These findings not only validate the effectiveness of our model but also highlight the potential of NLP-based techniques in revolutionizing computational protein design and bioengineering. The integration of bioinformatic supervision ensures that the generated sequences are not only plausible but also structurally viable, which is crucial for practical applications. 

Additional work in this direction can include conditional generation, where multiple properties can be targeted simultaneously within a single model. This advancement would enable the design of multifunctional proteins with combined therapeutic properties, optimizing the efficacy and application potential of synthetically designed peptides. 

\section{Data and Software Availability}
The necessary code and data used in this study can be accessed here:
\url{https://github.com/aayush-shah14/PeptideGPT}

\section{Supporting Information}

\subsection{Optimal Hyperparameters and Training Results}

The hyperparameters related to the fine-tuning process were carefully tuned to achieve the best loss on the validation dataset, the values are given in Table \ref{otmhp}. The loss is inversely proportional to the confidence of the model in predicting the next token given the previous tokens as a context. The number of epochs was kept to allow the loss curves to converge sufficiently. We can see that the hemolytic tasks achieved the lowest validation loss, followed by non-fouling and then solubility. 

\begin{table}[H]
\centering
\begin{tabular}{lcccc}
\hline
\textbf{Task}              & \textbf{Hemolytic} & \textbf{Non-Hemolytic} & \textbf{Non-Fouling} & \textbf{Soluble} \\ \hline
\rule{0pt}{1\normalbaselineskip}
\hspace{-1mm}\textbf{Learning rate} & $10^{-5}$          & $10^{-6}$               & $10^{-5}$            & $10^{-6}$        \\
\textbf{Epochs}        & 200                & 500                     & 200                  & 50               \\
\textbf{Validation Loss}     & 3.98               & 3.98                    & 4.62                 & 5.68             \\
\textbf{Train Loss}    & 2.88               & 3.94                    & 4.74                 & 5.86             \\ \hline
\end{tabular}
\caption{Optimal hyperparameters and training results}
\label{otmhp}
\end{table}

\subsection{Sampling Parameters}

Sampling from the trained model is influenced by parameters like the maximum length of returned sequences, the 'k' most likely tokens to consider for generation and the penalty for repeating tokens. We kept the repetition penalty and $top_k$ metric constant for all generation experiments since those were the best-performing parameters found for ProtGPT2. We generated 5000 sequences for each property, comprising 1000 sequences for five different seeds. The \% invalid proteins specify how many peptides, out of all the generated sequences, were eliminated due to lying outside the permissible convex hull of valid proteins. We also report the percentage of these valid protein sequences which scored a pLDDT value of above 70 through ESMFold to illustrate how many sequences possess a stable and ordered structure, which is essentially the final output of our approach. Non-hemolytic task had the highest percentage of proteins with pLDDT above 70, followed by hemolytic, non-fouling, and soluble. 

\begin{table}[H]
\centering
\begin{tabular}{lcccc}
\hline
                         & \textbf{Hemolytic} & \textbf{Non-Hemolytic} & \textbf{Non-Fouling} & \textbf{Soluble} \\ \hline
\textbf{Max. length}      & 10                 & 10                     & 5                    & 50               \\
\textbf{Top k}           & 950                & 950                    & 950                  & 950              \\
\textbf{Repetition Penalty} & 1.2              & 1.2                    & 1.2                  & 1.2              \\
\textbf{No. Generated Sequences} & 5000 & 5000 & 5000 & 5000 \\
\textbf{\% Invalid Proteins} & 6.60            & 12.04                  & 2.51                 & 4.65             \\
\textbf{\% pLDDT \textgreater 70} & 21.70      & 44.46                  & 9.52                 & 9.89             \\ \hline
\end{tabular}
\caption{Sampling parameters and results}
\label{tab:my_label}
\end{table}

\bibliography{reference}

\end{document}